\documentclass[letterpaper, 10 pt, conference]{URL-ieeeconf}
\IEEEoverridecommandlockouts    %
\usepackage{graphics}           
\usepackage{times}              
\usepackage{amsmath}            
\usepackage{amssymb}            
\usepackage{graphicx}
\usepackage{algorithm}
\usepackage[noend]{algpseudocode}
\usepackage{booktabs}
\usepackage{color}
\usepackage{multirow}
\usepackage{subcaption}
\usepackage{rotating}
\usepackage{cite}
\definecolor{instructioncolor}{rgb}{.5,.5,.5}
\usepackage{cuted}
\usepackage{caption}
\usepackage{makecell}        %
\usepackage{dsfont}          %
\usepackage{tabularx}        %
\usepackage[table]{xcolor}   %
\usepackage{caption}         %
\usepackage{placeins}

\def\secref#1{Section~\ref{#1}}
\def\figref#1{Fig.~\ref{#1}}
\def\tabref#1{Table~\ref{#1}}
\def\eqref#1{(\ref{#1})}

\def\vsfig{\vspace{-0.3cm}}
\def\vstab{\vspace{-0.3cm}}

\newcommand{\rom}[1]{\uppercase\expandafter{\romannumeral #1\relax}}

\makeatletter
\usepackage{xspace}
\DeclareRobustCommand\onedot{\futurelet\@let@token\@onedot}
\def\@onedot{\ifx\@let@token.\else.\null\fi\xspace}

\def\etal{{\textit{et al}}\onedot}
\makeatother

\usepackage{array}
\newcolumntype{L}[1]{>{\raggedright\let\newline\\\arraybackslash\hspace{0pt}}m{#1}}
\newcolumntype{C}[1]{>{\centering\let\newline\\\arraybackslash\hspace{0pt}}m{#1}}
\newcolumntype{R}[1]{>{\raggedleft\let\newline\\\arraybackslash\hspace{0pt}}m{#1}}

\newcommand{\norm}[1]{\lVert#1\lVert}

\renewcommand{\b}[1]{\mbox{\boldmath$#1$}}

\newcommand{\ba}{\b a}

\newcommand{\bn}{\b n}

\newcommand{\bp}{\b p}

\newcommand{\bz}{\b z}

\usepackage[table]{xcolor}
\usepackage{tabularx}
\newcolumntype{Y}{>{\centering\arraybackslash}X}

\definecolor{c1st}{RGB}{176,221,196}   %
\definecolor{c2nd}{RGB}{221,240,229}   %
\definecolor{refrow}{gray}{0.93}       %
\definecolor{tblgray}{gray}{0.55}

\newcommand{\fst}[1]{\cellcolor{c1st}\textbf{#1}}
\newcommand{\snd}[1]{\cellcolor{c2nd}#1}
\newcommand{\pz}{\phantom{0}}

\newcommand{\std}[1]{{\scriptsize\,$\pm$#1}}
\newcommand{\rs}{\rule[-0.5ex]{0pt}{2.6ex}}
\newcommand{\legendbox}[2]{{\setlength{\fboxsep}{1pt}\colorbox{#1}{#2}}}

\title{\LARGE \bf Which Taxels, Not How Many:\\ Tactile Sensor Placement for Dexterous Policies}

\author{Taeyun Kim$^{*}$, Jinsoo Shin, Taehwa Kim, Taeyoon Kim, and Jinyoung Choi%
\thanks{All authors are with NAVER LABS, Seongnam, Republic of Korea.}%
\thanks{$^{*}$Corresponding author: {\tt\small ktw1404@naverlabs.com}}%
}

\begin{document}
\maketitle
\thispagestyle{empty}
\pagestyle{empty}

\begin{abstract}
Policies for dexterous manipulation depend on design choices made before training: which sensors enter the observation, where they are placed, and how their readings are arranged for the encoder.
These choices are rarely examined.
For tactile sensing, simulation studies place taxels uniformly over the hand, and this placement has become a default without being tested.
We test this default under a single controlled protocol.
The protocol holds the teacher, the training procedure, and the candidate taxels fixed and varies only the sensor configuration and the representation.
We evaluate it on in-palm rotation and grasping with three robot hands.
A small set of taxels selected from contact statistics matches a far larger uniform set.
Taxels at rarely contacted sites are equivalent to no tactile input.
With all candidate taxels, their layout in the input tensor does not affect performance, even for a convolutional encoder.
Uniform placement therefore overestimates the number of taxels a task requires.
We distill these results into lessons for designing tactile policies and into a minimal set of design choices that a tactile policy study should report.
\end{abstract}
\vspace{-10pt}

\section{Introduction}
\label{sec:intro}

Humans adjust their grasp to the physical properties of everyday objects through tactile sensing~\cite{johansson2009nrn}.
In dexterous robotic manipulation, vision is often insufficient for contact-rich tasks, which involve occlusions and subtle visual changes~\cite{kim2026arxiv}. Tactile sensing allows robot hands to regulate grip force~\cite{zhao2026arxiv,kim2026arxiv}, manipulate objects hidden from view~\cite{kim2026arxiv,yin2023rss}, and infer object properties~\cite{qi2023rotateit}.

However, tactile signals differ in nature from visual and proprioceptive signals.
They are sparse and non-smooth, since they are present only during contact~\cite{miller2025neurips}.
These properties may lead to imbalanced training data and unstable learning~\cite{miller2025neurips}.
The articulated geometry of the hand also makes these signals spatially irregular, with contacts unevenly distributed across its regions and taxels placed on curved surfaces rather than on a regular grid~\cite{miller2025neurips,albini2025arxiv}.

How much spatial detail a policy needs from touch depends on the task.
In antipodal grasping, spatially resolved touch adds nothing beyond the net force on each finger~\cite{zhang2025arxiv}.
For in-hand rotation, in contrast, contact locations substantially improve performance~\cite{qi2023rotateit}.
Where a policy will make contact is unknown before training.
Simulation studies therefore place taxels uniformly over the hand by default~\cite{chung2026arxiv}, and our candidate set follows the same practice.

This default, however, is an assumption rather than a measured requirement.
A taxel that is never contacted returns only zeros.
It still consumes observation dimensions, simulation time, and, on a real hand, wiring and readout bandwidth.
Both placement and the encoder are chosen before any tactile policy is trained.
These choices differ across prior studies, so their effects are difficult to compare~\cite{miller2025neurips}.

Our goal is therefore to compare these design choices under a single controlled protocol.
To this end, we (1) fix the teacher, the training procedure, and the candidate taxels, and vary only the sensor configuration and the representation; (2) examine these choices on in-palm rotation with one hand, and test the placement results on a grasping task with two other hands; and (3) derive lessons for designing tactile policies.
From these results we draw three lessons.
First, which taxels are sensed matters more than how many.
Second, taxels at rarely contacted sites are equivalent to no touch.
Third, the encoder needs to preserve which taxel reports the contact, not its position in the input tensor.

\section{Related Work}
\label{sec:related}

\subsection{Tactile Representation for Policy Learning}
Sequence models assume an ordered sequence of tokens, and convolutional networks assume a regular grid of pixels.
Taxels on a robot hand satisfy neither assumption, since they lie on the curved surfaces of articulated links.
Tactile readings must be mapped to an input representation before a policy can process them, and this mapping determines what information is encoded~\cite{albini2025arxiv}.
Tactile images assume a planar sensor array, whereas mesh-based representations encode the adjacency between taxels on curved surfaces~\cite{albini2021ral}.
Dexterous manipulation policies pair convolutional encoders with grid inputs and MLP or LSTM encoders with grid-free inputs~\cite{chung2026arxiv}, and graph neural networks have been applied as well~\cite{yang2023ral}.
Chung \etal~\cite{chung2026arxiv} compare observation types with the tensor layout held fixed, and find per-taxel force and torque the most consistently useful.
Holding this signal fixed, we isolate the effects of the tensor layout and the encoder.

\subsection{Tactile Sensor Configuration for Efficient Sensing}
The number and placement of tactile sensors determine what information a policy receives.
Prior work has examined how sensor resolution affects learning efficiency and performance~\cite{melnik2021frontiers}, reduced the sensor count while retaining performance~\cite{guo2024arxiv}, and identified which regions of the palm and fingers contribute to manipulation~\cite{almeida2025arxiv}.
Even at a fixed sensor count, placement substantially changes learning outcomes, and this effect depends in part on the hand model and the learning algorithm~\cite{birtalan2026arxiv}.
Full-hand coverage outperforms fingertip placement on in-palm rotation~\cite{chung2026arxiv}, but the two differ in sensor count, and the taxels are sampled uniformly over the hand.
We fix both the candidate set and the sensor count and vary only the selection rule.
\clearpage

\newcommand{\taskname}[1]{\textbf{\textit{#1}}}
\newcommand{\methodname}[1]{\textbf{\textit{#1}}}

\section{Study Design}
\label{sec:design}
\subsection{Research Questions}
To examine how design choices in the learning pipeline affect policy performance, this study addresses the following two research questions.
\begin{itemize}
  \item \textbf{RQ1 (Sensor configuration):} How do the number and placement of taxels affect policy performance, and does the effect of placement depend on the taxel budget?
  \item \textbf{RQ2 (Representation):} Given the same taxel signal, how do the tensor layout and encoder affect policy performance?
\end{itemize}
To answer these questions, we conduct experiments across various robot hands and tasks with multiple random seeds.
\vspace{-1pt}

\subsection{Tasks and Embodiments}
\begin{figure}[H]
  \centering
  \includegraphics[width=\columnwidth]{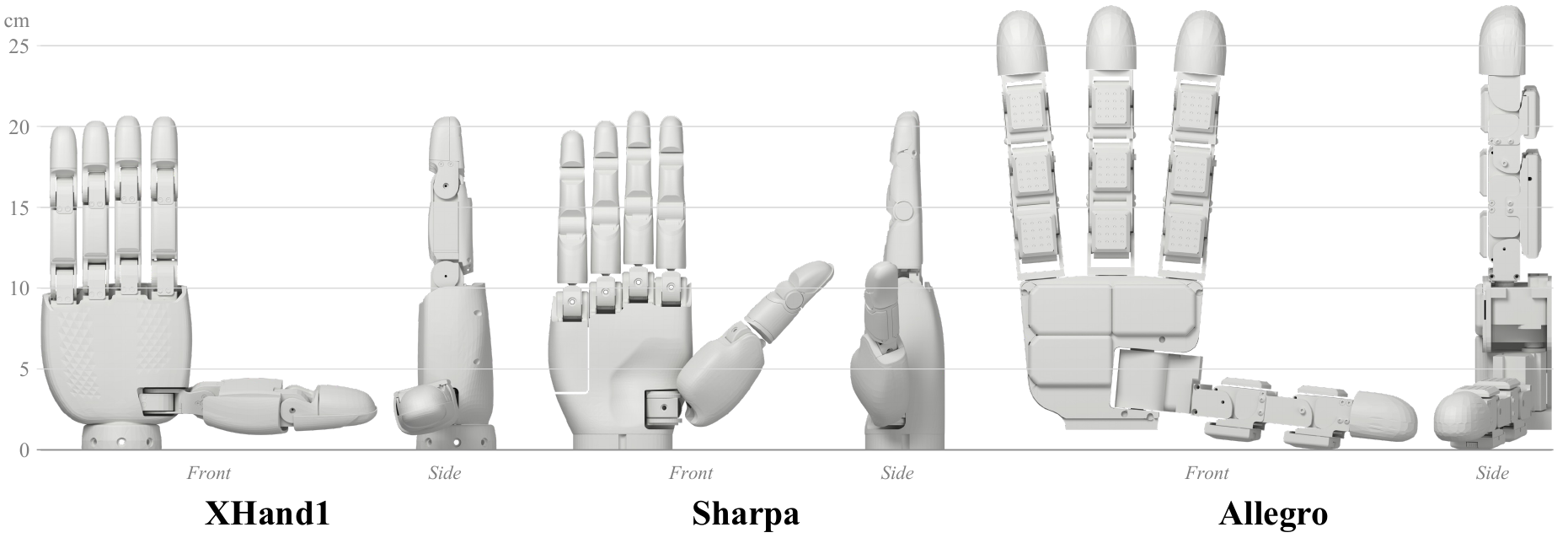}
  \caption{\textbf{Three dexterous hands at a common scale.}
  Front and side views of XHand1 (12 DoF), Sharpa (22 DoF), and Allegro (16 DoF).}
  \label{fig:hands}
  \vspace{-10pt}
\end{figure}

We evaluate two dexterous manipulation tasks on the hands in \figref{fig:hands}, all simulated in Genesis~\cite{genesis}.
The objects used in each task are listed in Appendix~\ref{app:tasks}.

\taskname{In-Palm Rotation (XHand1).}
The hand must maintain an object in the palm while rotating it in successive quarter turns in one direction about a fixed horizontal axis~\cite{chung2026arxiv}.
Only the thumb, the middle finger, and the palm are used in this task.

\taskname{Grasp-and-Lift (Sharpa, Allegro).}
The hand must grasp an object from a table, lift it, and carry it to a target position above the table~\cite{chung2026arxiv}.
While the object is held in the air, external disturbances are applied, and the policy must maintain the grasp.

\subsection{Teacher--Student Distillation}
Each dexterous task is formulated as a partially observable Markov decision process (POMDP), defined by the tuple $(\mathcal{S}, \mathcal{O}, \mathcal{A}, P, \mathcal{R}, \gamma)$, which specifies the state, observation, and action spaces, transition dynamics, reward function, and discount factor, respectively.
Following prior work~\cite{qi2022hora, qi2023rotateit}, policy training consists of two stages.
First, a teacher policy $\pi_T : \mathcal{S} \to \mathcal{A}$ is trained with proximal policy optimization (PPO)~\cite{schulman2017ppo} on privileged object properties that are not directly observable.
The teacher receives no tactile observations, so its contacts do not depend on any taxel placement.
Next, a student policy $\pi_S$ is distilled from $\pi_T$ using DAgger~\cite{ross2011dagger}.
The student receives only observations available at deployment: the history of joint states, the goal position, and the tactile observation $o^{\mathrm{tac}}_t$.
It is optimized with the following objective:
\begin{equation}
\min_{\pi_S} \; \mathbb{E}\left[ \left\| \pi_S(o_{1:t}) - \pi_T(s_t) \right\|_2^2 \right],
\label{eq:distill}
\end{equation}
where the expectation is taken over the states visited by $\pi_S$.
All instances of $\pi_S$ are distilled from the same $\pi_T$ and differ only in the tactile observation $o^{\mathrm{tac}}_t$ and its encoder.
Further training details are provided in Appendix~\ref{app:training}.

\subsection{Tactile Sensor Configuration}
\begin{figure}[H]
  \centering
  \captionsetup{font=footnotesize}
  \includegraphics[width=\columnwidth]{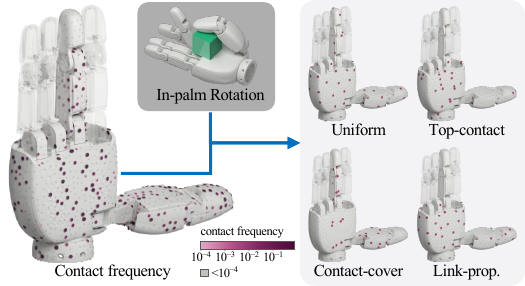}
   \caption{\textbf{Taxel placement on the XHand1 for In-Palm Rotation.}
  Contact frequency of $\pi_T$ at each candidate taxel and the taxels selected by each placement method (shown for $B{=}100$).
  Faded fingers are not used.}                           %
  \label{fig:placement}
  \vsfig
\end{figure}
Following Chung \etal~\cite{chung2026arxiv}, we sample $N = 2000$ candidate taxels uniformly on the surface of the links used in each task by Poisson-disk sampling (\figref{fig:placement}).
Each taxel $i$ measures the force and torque in the frame of its link, denoted by $x_{i,t} \in \mathbb{R}^6$ at timestep $t$.
A placement method selects $B$ of the $N$ candidates, where $B$ denotes the taxel budget and ranges from 10 to 300.
The tactile observation $o^{\mathrm{tac}}_t$ consists of the measurements $x_{i,t}$ of the selected taxels.
For each placement method, the selected taxels are nested across budgets, so a larger budget only adds taxels.
For \taskname{In-Palm Rotation}, unselected taxels are zero-padded so that the input dimension remains constant across all budgets and placement methods.

To isolate the effect of placement, we select all placements from the same candidate set.
Every task-informed method relies on the same contact statistics of $\pi_T$.
The methods differ therefore only in the criterion by which taxels are selected; we compare four placement methods and one control (\figref{fig:placement}).
Let $\mathcal{E}_i$ denote the set of timesteps at which taxel $i$ reports a non-zero reading in rollouts of $\pi_T$.
A contact event is a timestep at which a link is in contact.
\begin{itemize}
    \item \methodname{Top-contact} selects the $B$ taxels with the largest $|\mathcal{E}_i|$.
  \item \methodname{Uniform} allocates taxels to each link in proportion to its number of candidates, without any task information.
  \item \methodname{Contact-cover} greedily adds the taxel that detects the most contact events not yet detected by a selected taxel on the same link.
  \item \methodname{Link-proportional} allocates taxels to each link in proportion to its share of contacts.
  \item \methodname{Least-contact} selects the $B$ contacted taxels with the smallest $|\mathcal{E}_i|$ and serves as a control.
\end{itemize}
\methodname{Uniform} and \methodname{Link-proportional} spread the allocated taxels within each link by farthest point sampling. For In-Palm Rotation, \methodname{Top-contact}, \methodname{Contact-cover}, and \methodname{Least-contact} then assign at least one taxel to every link.
As a result, \methodname{Top-contact} concentrates taxels on frequently contacted spots, whereas \methodname{Contact-cover} and \methodname{Link-proportional} distribute them across links. As a reference without spatial resolution, we also consider \methodname{Per-link force}, which replaces the taxel measurements with the net contact force on each link in the same frame.

\subsection{Tactile Representation}
\vspace{-10pt}
\begin{figure}[H]
  \centering
  \includegraphics[width=\columnwidth]{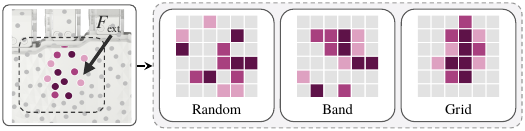}
  \caption{\textbf{Tensor layouts of the selected taxels.}
  An example contact caused by an external force $F_{\mathrm{ext}}$ on the palm is mapped into the tensor under each layout.}
  \vspace{-10pt}
  \label{fig:layout}
\end{figure}
\label{sec:representation}
Convolutional encoders assume that adjacent input cells correspond to neighboring measurements.
Shuffling pixel positions violates this assumption and degrades convolutional networks, whereas ViTs are unaffected~\cite{jelassi2022nips}.
Unlike image pixels, taxels lie on curved surfaces without a predefined grid~\cite{albini2025arxiv}.
Prior tactile policies nonetheless pair grid inputs with convolutional encoders~\cite{chung2026arxiv}.
The effect of the tensor layout may therefore depend on the encoder.
To isolate this interaction, we compare representations with all $N$ candidates, where placement does not intervene, and vary only the layout and the encoder (\figref{fig:layout}).
Each link occupies a fixed block of cells, and the three layouts differ only in which taxel is assigned to which cell.
\begin{itemize}
  \item \methodname{Random} fills the cells in a fixed random order.
  \item \methodname{Band} fills the cells in a serpentine order along each link.
  \item \methodname{Grid} assigns each taxel to the cell nearest its surface position.
\end{itemize}

We compare three encoders that differ in how they treat the identity and the adjacency of taxels.
A convolutional network reads each cell by its position and relies on cell adjacency.
An LSTM reads each cell through its own input weight and retains the identity of each taxel without any adjacency assumption.
A link-wise MLP averages over the cells of each link and over links.
All three students share the proprioceptive encoder and the policy head, and differ only in the tactile encoder (Appendix~\ref{app:encoders}).

\section{Results}
\label{sec:results}

\subsection{Evaluation Metric}
\label{sec:metric}
Each policy is evaluated over 1024 episodes, and results are averaged over multiple random seeds.
Since the results of policy learning can vary with the point at which training is stopped~\cite{mandlekar2021corl}, all policies are evaluated at the same point in training.
For \taskname{In-Palm Rotation}, we report the success rate.
For \taskname{Grasp-and-Lift}, we report the held fraction under external disturbances.
As references, we report the performance of $\pi_T$ and of $\pi_S$ without tactile observations.
Further details of the metrics and the evaluation protocol are provided in Appendix~\ref{app:metric}.

\begin{table}[t]
  \centering
  \footnotesize
  \setlength{\tabcolsep}{2pt}
  \renewcommand{\arraystretch}{0.75}
  \captionsetup{font=footnotesize}
  \caption{\textbf{Success rate (\%) across placement methods and taxel budgets.}
  \taskname{In-Palm Rotation} on the XHand1 with the LSTM encoder. Mean $\pm$ std over three seeds.}
  \label{tab:placement}
  \vspace{-1mm}
  \begin{tabularx}{\columnwidth}{l >{\hsize=0.96\hsize}Y >{\hsize=1.04\hsize}Y >{\hsize=1.12\hsize}Y >{\hsize=0.88\hsize}Y}
  \toprule
  & \multicolumn{1}{c}{\textit{Task-agnostic}}
  & \multicolumn{3}{c}{\textit{Task-informed}} \\
  \cmidrule(lr){2-2}
  \cmidrule(l){3-5}
  \rs Budget ($B$) & Uniform & Top-contact & Contact-cover & Link-prop. \\
  \midrule
  \rowcolor{refrow}\rs\textit{No tactile}     & \multicolumn{4}{c}{\cellcolor{refrow}\pz0.0\std{0.0}} \\
  \rowcolor{refrow}\rs\textit{Least-contact}  & \multicolumn{4}{c}{\cellcolor{refrow}\pz0.0\std{0.0}} \\
  \rowcolor{refrow}\rs\textit{Per-link force} & \multicolumn{4}{c}{\cellcolor{refrow}\pz4.4\std{1.0}} \\
  \specialrule{0.4pt}{0pt}{1.2pt}
  \specialrule{0.4pt}{0pt}{0pt}
  \rs10  & \pz0.0\std{0.0} & \fst{41.4\std{0.6}} & \snd{38.7\std{2.4}} & \pz0.0\std{0.0} \\
  \rs15  & \pz0.0\std{0.0} & \fst{42.5\std{7.6}} & \snd{42.3\std{1.0}} & \pz0.0\std{0.0} \\
  \rs25  & 12.4\std{1.3}   & \fst{50.5\std{1.3}} & \snd{50.0\std{1.8}} & \pz8.6\std{3.1} \\
  \rs50  & 27.5\std{1.5}   & \fst{53.3\std{3.7}} & \snd{52.8\std{5.6}} & 40.8\std{4.3} \\
  \rs75  & 35.5\std{6.7}   & \fst{59.5\std{1.2}} & \snd{54.8\std{1.8}} & 47.9\std{3.5} \\
  \rs100 & 51.4\std{2.3}   & \fst{56.6\std{2.9}} & \snd{55.7\std{3.8}} & 48.8\std{0.9} \\
  \rs300 & 56.2\std{3.6}   & \snd{58.0\std{2.8}} & \fst{59.4\std{2.8}} & 56.9\std{1.3} \\
  \specialrule{0.4pt}{0pt}{1.2pt}
  \specialrule{0.4pt}{0pt}{0pt}
  \rowcolor{refrow}\rs\textit{All (2000)}      & \multicolumn{4}{c}{\cellcolor{refrow}61.5\std{0.3}} \\
  \rowcolor{refrow}\rs\textit{Teacher $\pi_T$} & \multicolumn{4}{c}{\cellcolor{refrow}93.8} \\
  \bottomrule
  \end{tabularx}

  \par\vspace{2pt}
  {\footnotesize\raggedright
  \textbullet\; \legendbox{c1st}{\textbf{Best}} and \legendbox{c2nd}{second-best} per budget are highlighted.
  \textit{Least-contact}: taxels at the least contacted spots ($B{=}50$ and $100$).\par}
  \vstab
\end{table}

\subsection{Effect of Sensor Configuration}
\label{sec:sensor}

Table~\ref{tab:placement} shows the success rate of $\pi_S$ across budgets and placement methods.
Policy performance depends more on the selection of taxels than on their number.
A small set of taxels selected from contact statistics reaches the success rate of a far larger uniform set.
\methodname{Top-contact} and \methodname{Contact-cover} place taxels where $\pi_T$ makes contact and therefore detect most of its contact events with few taxels (Appendix~\ref{app:analysis}).

Taxels placed where contact rarely occurs carry no information.
\methodname{Least-contact}, which places taxels at the least contacted sites, never succeeds, the same as $\pi_S$ without tactile observations.
\methodname{Uniform} fails in the same way at the lowest budgets, where the selected taxels detect few of the contact events of $\pi_T$ (Appendix~\ref{app:analysis}).
Uniform placement appears to require many taxels because most of them return zeros, not because the task requires that many.

Tactile observations drive the rotation in \taskname{In-Palm Rotation}.
Without them, $\pi_S$ holds the object in most episodes but does not rotate it.
\methodname{Per-link force}, which has no spatial resolution, offers little improvement, so spatially resolved contact information is required.
At the lowest budgets, $\pi_S$ drops the object more often than without tactile observations, which indicates that it attempts the rotation with insufficient contact information.
As the budget grows, both drops and under-rotation decrease (Appendix~\ref{app:analysis}).

The success rate of the task-informed methods saturates from intermediate budgets onward, and \methodname{Uniform} catches up at the largest budget, where all methods converge to a comparable success rate close to that of the full candidate set.
For \methodname{Uniform}, the taxels selected at higher budgets lie closer together, and the fraction of a taxel's contacts also detected by its nearest neighbor increases substantially (Appendix~\ref{app:analysis}).
The information gained from additional taxels therefore diminishes.
This is consistent with the finding of Chung \etal~that coverage matters more than resolution~\cite{chung2026arxiv}.
\methodname{Link-proportional} moves taxels from rarely contacted links to the thumb tip (\figref{fig:placement}).
It improves on \methodname{Uniform} at intermediate budgets, matches it at $B{\geq}100$, and, like \methodname{Uniform}, performs poorly at $B{\leq}25$.
Even with all taxels, a gap from $\pi_T$ remains.

\begin{table}[t]
  \centering
  \footnotesize
  \setlength{\tabcolsep}{2pt}
  \renewcommand{\arraystretch}{0.75}
  \captionsetup{font=footnotesize}
  \caption{\textbf{Held fraction (\%) across taxel budgets and placement methods.}
  \taskname{Grasp-and-Lift} on the Sharpa and the Allegro with the LSTM encoder.}
  \label{tab:grasplift}
  \vspace{-1mm}
  \begin{tabularx}{\columnwidth}{l >{\hsize=0.96\hsize}Y >{\hsize=1.04\hsize}Y >{\hsize=1.12\hsize}Y >{\hsize=0.88\hsize}Y}
  \toprule
  & \multicolumn{1}{c}{\textit{Task-agnostic}}
  & \multicolumn{3}{c}{\textit{Task-informed}} \\
  \cmidrule(lr){2-2}
  \cmidrule(l){3-5}
  \rs Budget ($B$) & Uniform & Top-contact & Contact-cover & Link-prop. \\
  \midrule
  \multicolumn{5}{@{}l}{\rs\textbf{\textit{Sharpa}}} \\
  \rowcolor{refrow}\rs\textit{No tactile} & \multicolumn{4}{c}{\cellcolor{refrow}\pz9.4\std{0.7}} \\
  \specialrule{0.4pt}{0pt}{1.2pt}
  \specialrule{0.4pt}{0pt}{0pt}
  \rs25  & 15.9\std{0.7}  & \snd{22.3\std{6.7}}  & \fst{27.9\std{8.5}}  & 21.2\std{3.9} \\
  \rs50  & 24.0\std{0.8}  & \snd{29.1\std{12.5}} & \fst{38.9\std{15.7}} & 26.6\std{5.5} \\
  \rs100 & 25.6\std{3.4}  & 24.3\std{1.3}        & \fst{32.0\std{5.4}}  & \snd{26.9\std{2.8}} \\
  \rs300 & 42.6\std{2.1}    & 27.0\std{3.7}         & \snd{57.0\std{4.7}}    & \fst{59.8\std{2.2}} \\
  \specialrule{0.4pt}{0pt}{1.2pt}
  \specialrule{0.4pt}{0pt}{0pt}
  \rowcolor{refrow}\rs\textit{Teacher $\pi_T$} & \multicolumn{4}{c}{\cellcolor{refrow}96.0} \\
  \addlinespace[3pt]
  \specialrule{0.8pt}{0pt}{2pt}
  \multicolumn{5}{@{}l}{\rs\textbf{\textit{Allegro}}} \\
  \rowcolor{refrow}\rs\textit{No tactile} & \multicolumn{4}{c}{\cellcolor{refrow}22.9\std{2.1}} \\
  \specialrule{0.4pt}{0pt}{1.2pt}
  \specialrule{0.4pt}{0pt}{0pt}
  \rs25  & 39.5\std{4.6}        & 22.8\std{3.5}       & \fst{45.5\std{3.1}}   & \snd{40.0\std{9.2}} \\
  \rs50  & 43.3\std{3.4}        & 38.2\std{1.0}       & \snd{45.9\std{8.8}} & \fst{46.1\std{3.1}} \\
  \rs100 & 59.0\std{16.1}       & \snd{60.4\std{3.2}} & \fst{61.3\std{0.7}} & 52.9\std{7.1} \\
  \rs300 & 57.0\std{11.9}        & \snd{57.2\std{1.6}} & \fst{72.2\std{2.1}}   & 54.2\std{4.2} \\
  \specialrule{0.4pt}{0pt}{1.2pt}
  \specialrule{0.4pt}{0pt}{0pt}
  \rowcolor{refrow}\rs\textit{Teacher $\pi_T$} & \multicolumn{4}{c}{\cellcolor{refrow}93.0} \\
  \bottomrule
  \end{tabularx}

    \par\vspace{2pt}
  {\footnotesize\raggedright
  \textbullet\; \legendbox{c1st}{\textbf{Best}} and \legendbox{c2nd}{second-best} per budget and hand are highlighted.\par}
  \vstab
  \vspace{-10pt}
\end{table}

Table~\ref{tab:grasplift} shows the results on \taskname{Grasp-and-Lift}.
In this task, every $\pi_S$ makes first contact within 1\,s and lifts the object in almost every episode, including the one without tactile observations.
The benefit of tactile observations therefore arises in holding the object under disturbances.
The two hands differ in held fraction even without tactile observations.
The effect of placement follows the opposite trend to \taskname{In-Palm Rotation}: except for \methodname{Top-contact}, the methods differ little at low budgets and diverge at the largest budget.
At that budget, \methodname{Contact-cover} exceeds \methodname{Uniform} on both hands, and \methodname{Top-contact} falls well below it on the Sharpa.
Concentrating taxels on the most contacted sites provides no advantage.
\methodname{Top-contact} does not consistently outperform \methodname{Uniform}, and at the lowest budget on the Allegro it performs no better than the policy without tactile observations.
The contacts of $\pi_T$ spread over more links in this task than in \taskname{In-Palm Rotation} (Appendix~\ref{app:analysis}), so covering the contacted links matters more than concentrating taxels on one of them.

\subsection{Effect of Representation}
\label{sec:representation}

Table~\ref{tab:representation} shows the success rate across encoders and tensor layouts with all candidate taxels.
With the same taxels and the same signal, the encoder has a much larger effect than the tensor layout.
For the CNN, the three layouts yield comparable success rates, although \methodname{Grid} preserves surface neighbors and \methodname{Random} destroys them.
This differs from image classification, where shuffling pixels degrades convolutional networks~\cite{jelassi2022nips}.

For the LSTM, which reads each cell through its own input weight, a fixed permutation of taxels does not limit what it can represent, and \methodname{Random} and \methodname{Grid} perform comparably.
Policy performance depends on the identity of the taxels, whereas their order in the tensor has no effect.
The link-wise MLP falls far below both.
It shares the recurrent module of the CNN and differs only in the spatial module (Appendix~\ref{app:encoders}).
This module averages over the taxels of each link and discards their identity.
This is consistent with the need for spatially resolved contact information in \secref{sec:sensor}.
\begin{table}[t]
  \centering
  \footnotesize
  \setlength{\tabcolsep}{7pt}
  \renewcommand{\arraystretch}{0.75}
  \captionsetup{font=footnotesize}
  \caption{\textbf{Success rate (\%) across encoders and tensor layouts.}
  \taskname{In-Palm Rotation} on the XHand1 with all 2000 taxels.}
  \label{tab:representation}
  \vspace{-1mm}
  \begin{tabular}{l ccc}
  \toprule
  & \multicolumn{3}{c}{\textit{Tensor layout}} \\
  \cmidrule(l){2-4}
  \rs Encoder & Random & Band & Grid \\
  \specialrule{0.4pt}{0pt}{1.2pt}
  \specialrule{0.4pt}{0pt}{0pt}
  
    \rs CNN           & 61.4\std{0.6} & 64.3\std{1.3} & 61.3\std{1.1} \\
    \rs LSTM          & 60.8\std{3.9} & 58.0\std{3.2} & 61.5\std{0.3} \\
    \rs Link-wise MLP & 33.2\std{8.6} & 25.2\std{3.3} &18.6\std{1.1}  \\

  \bottomrule
  \end{tabular}
  \vstab
  \vspace{-10pt}
\end{table}

\subsection{Training Resource}
\label{sec:cost}
\vspace{-10pt}
\begin{table}[H]
  \centering
  \footnotesize
  \setlength{\tabcolsep}{5pt}
  \renewcommand{\arraystretch}{0.75}
  \captionsetup{font=footnotesize}
  \caption{\textbf{Training cost per tactile observation size.}
    \taskname{In-Palm Rotation} with 2048 environments on NVIDIA H100 GPU. Speed-up is relative to the full observation used in this study.}
  \label{tab:cost}
  \vspace{-1mm}
  \begin{tabular}{l r r r}
  \toprule
  \rs Tactile observation & Env-steps/s & Speed-up & VRAM \\
  \specialrule{0.4pt}{0pt}{1.2pt}
  \specialrule{0.4pt}{0pt}{0pt}
  \rs None            & 42{,}810   & 7.6$\times$ & \pz1.5\,GB \\
  \rs $B{=}25$        & 32{,}233   & 5.7$\times$ & \pz1.6\,GB \\
  \rs $B{=}100$       & 29{,}503   & 5.3$\times$ & \pz1.9\,GB \\
\rs $B{=}300$       & 23{,}817   & 4.2$\times$ & \pz2.5\,GB \\
  \rs $B{=}2000$      & \pz8{,}411 & 1.5$\times$ & \pz8.1\,GB \\
  \rowcolor{refrow}\rs Full (ours) & \pz5{,}610 & 1.0$\times$ & 12.4\,GB \\
  \bottomrule
  \end{tabular}
  \vstab
\end{table}

Table~\ref{tab:cost} shows the training throughput and VRAM for each size of the tactile observation.
At the budget where the success rate saturates, simulating only the selected taxels runs about five times faster than the full observation, at a fraction of the memory.
The two rows with all 2000 taxels differ only in the zero-padded cells.
Padding alone reduces the throughput by a third and increases the memory by half.
All policies in this paper are trained with the full observation.

\section{Conclusion}
\label{sec:conclusion}
We compared tactile design choices for dexterous policies under a single controlled protocol that varies only the sensor configuration and the representation.
A few taxels placed where $\pi_T$ makes contact matched a far larger set placed uniformly.
Taxels placed where contact rarely occurred provided no usable information.
With all candidate taxels, their layout in the input tensor did not affect performance even for a convolutional encoder.

These results share a single explanation.
Tactile signals carry information only where contact occurs.
Uniform placement ignores this property and therefore overestimates the number of taxels a task requires.

These lessons come from one hand and one task in which contact concentrates on a few links.
On \taskname{Grasp-and-Lift}, where contact spreads over more links, the effect of placement followed a different trend.
The two tasks nonetheless point to the same conclusion.
The question for tactile sensing is not how many taxels to place but which ones.
The answer can be read from the contact statistics of $\pi_T$ before any tactile policy is trained.
Uniform placement thus changes from a hidden default into a measurable design choice.

The contact statistics came from $\pi_T$, but contact locations recorded from rollouts of any policy or from human demonstrations can select the taxels in the same way.
All results were obtained in simulation, and the effect of sensor noise and drift remains untested.
Appendix~\ref{app:checklist} lists the design choices that studies of tactile policies should consider.
\clearpage

\bibliographystyle{URL-IEEEtrans}

\bibliography{URL-bib}
\appendix
\subsection{Reporting Checklist for Tactile Observations}
\label{app:checklist}
\vspace{-10pt}
\begin{table}[H]
  \centering
  \footnotesize
  \setlength{\tabcolsep}{3pt}
  \renewcommand{\arraystretch}{0.9}
  \captionsetup{font=footnotesize}
  \caption{\textbf{Reporting checklist for tactile observations in policy learning.}
  Settings refer to \taskname{In-Palm Rotation} on the XHand1 unless noted. Items varied in this study are marked with $\dagger$.}
  \label{tab:checklist}
  \vspace{-1mm}
  \begin{tabular}{@{}>{\raggedright\arraybackslash\leftskip=1em}p{0.26\columnwidth}>{\raggedright\arraybackslash}p{0.66\columnwidth}@{}}
  \toprule
  \multicolumn{1}{@{}l}{\rs Design choice} & Setting \\
  \specialrule{0.4pt}{0pt}{1.2pt}
  \specialrule{0.4pt}{0pt}{0pt}
  \rowcolor{refrow}\multicolumn{2}{@{}l}{\rs\textit{Sensing}} \\
  \rs Candidate set       & $N{=}2000$, Poisson-disk sampling on the used links \\
  \rs Taxel signal        & 6-D force and torque in the link frame \\
  \rowcolor{refrow}\multicolumn{2}{@{}l}{\rs\textit{Selection (RQ1)}} \\
  \rs Budget$^\dagger$         & $B\in\{10, 15, 25, 50, 75, 100, 300, 2000\}$\\
  \rs Placement rule$^\dagger$ & Uniform, Top-contact, Contact-cover, \mbox{Link-proportional}, and Least-contact \\
  \rs Contact statistics  & rollouts of $\pi_T$ (no tactile input) \\
  \rowcolor{refrow}\multicolumn{2}{@{}l}{\rs\textit{Representation (RQ2)}} \\
  \rs Tensor size         & fixed at $N$ taxels, with zero padding~(Table~\ref{tab:cost}) \\  \rs Ordering$^\dagger$       & Random, Band, Grid \\
  \rs Encoder$^\dagger$        & LSTM, CNN, Link-wise MLP \\
  \rowcolor{refrow}\multicolumn{2}{@{}l}{\rs\textit{Evaluation}} \\
  \rs Protocol            & 1024 episodes, mean and standard deviation \\
  \rs Metric              & success rate (\taskname{In-Palm Rotation})\newline held fraction under disturbance (\taskname{Grasp-and-Lift}) \\
  \bottomrule  
  
  \end{tabular}
\end{table}

\vspace{-10pt}
A tactile observation is determined by several choices made before any policy is trained.
Prior studies make these choices differently, and some leave them unreported.
\tabref{tab:checklist} organizes these choices into four groups and lists the setting used in this study for each.

\FloatBarrier
\subsection{Tasks and Objects} 
\label{app:tasks}
\begin{figure}[H]
  \centering
  \includegraphics[width=\columnwidth]{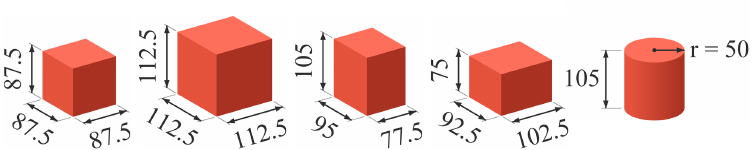}
  \caption{\textbf{Objects used in the two tasks.}~The object set consists of 15 boxes and 2 cylinders, five of which are shown at the Allegro scale. Dimensions are given in mm.}
\label{fig:task_objects}
  \vspace{-10pt}
\end{figure}

\begin{figure*}[!t]
  \centering
  \includegraphics[width=\textwidth]{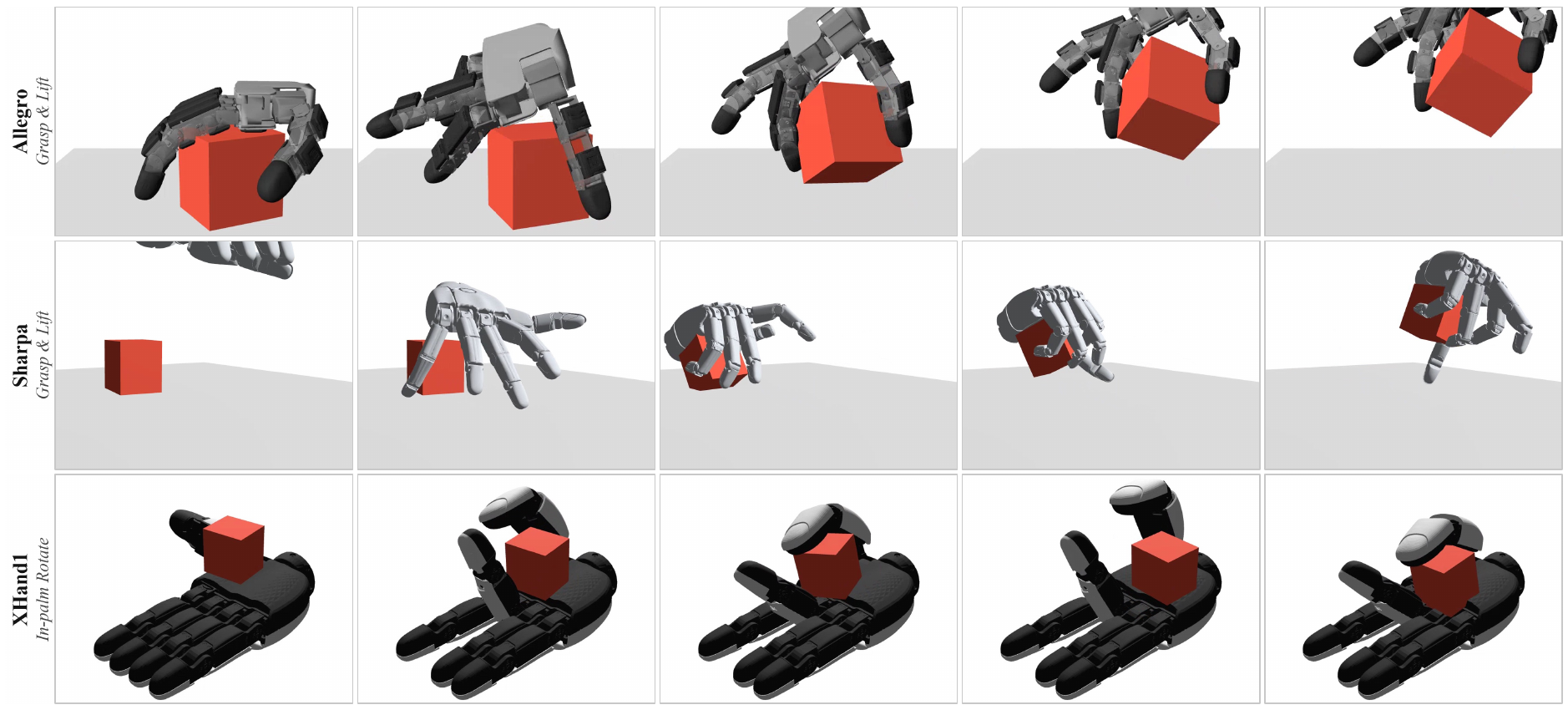}
  \caption{\textbf{Teacher rollouts on the three hands.}~Each row shows frames sampled at 2\,s intervals from a single episode of $\pi_T$. The Allegro and the Sharpa are shown on \taskname{Grasp-and-Lift}, and the XHand1 on \taskname{In-Palm Rotation}.}
\label{fig:teacher_rollouts}
  \vspace{-10pt}
\end{figure*}

The three hands share one object set, scaled to the size of each hand.
\figref{fig:task_objects} shows the objects, and \figref{fig:teacher_rollouts} shows rollouts of $\pi_T$ on each task.
Each parallel environment is assigned one object when the scene is built, and the object is held fixed across episodes.
Environments are divided evenly among the objects, so every object appears in the same number of environments in both training and evaluation.

In \taskname{In-Palm Rotation}, the object starts resting on the palm.
An episode terminates when the object falls from the hand.

In \taskname{Grasp-and-Lift}, the object starts on the table and the hand above it.
External disturbances are applied only after the object has been lifted; they are specified in Appendix~\ref{app:metric}.
An episode terminates when the object falls below the table.

Object mass and friction, the initial pose of the object, and the actuator parameters are randomized at every reset, and $\pi_S$ uses the same randomization as $\pi_T$.

\FloatBarrier
\subsection{Teacher and Student} 
\label{app:training}

Teachers and students are trained following the procedure in \secref{sec:design}.
Each teacher is optimized with PPO on the rewards below and then held fixed, and each student is distilled from its teacher with DAgger ($\beta{=}0$) and the executed-action MSE loss.
\tabref{table:reward_inpalm} and \tabref{table:reward_grasp} list the reward terms and weights of each teacher.
Every reward is the sum of a task reward and a regularization reward,
\begin{equation}
\mathcal{R} = \mathcal{R}_{\text{task}} + \mathcal{R}_{\text{reg}},
\label{eq:reward}
\end{equation}
where each component is a weighted sum $\sum_k w_k r_k$ of the terms in the corresponding table, and $[x]_+ = \max(0, x)$.

For In-Palm Rotation, Goal Reached and Rotation Progress drive the object toward the commanded orientation, where $\theta_t$ denotes the orientation error to the goal and $\Delta\theta_t$ the rotation progress about the command axis.
Off-Axis Orientation penalizes the component $\theta^{\perp}_t$ of this error orthogonal to the command axis.
Fingertip Proximity encourages contact through the smallest fingertip-to-object distance $d_{\min,t}$, Object Displacement penalizes deviation from the nominal object position $\bp^{\ast}$, and Contact Force penalizes excessive contact force $\mathbf{f}_t$ on the object.

For Grasp-and-Lift, Sustained Lift rewards the object height $h_t$ above the support plane, and Goal Tracking rewards proximity of the object to the commanded position $\bp^{\text{goal}}$.
Fingertip Proximity averages over the distances $d_{i,t}$ from the $N_\text{finger}$ fingertips to the object, and Palm Orientation penalizes a palm normal $\hat{\bn}_t$ that faces downward.
Sharpa and Allegro share the same terms and weights.

The rewards are used only for teacher training and for monitoring student runs.
\tabref{table:training} lists the simulation, PPO, and distillation hyperparameters.
Student runs retain the teacher's reward, termination, and randomization settings and differ only in the number of parallel environments.

\begin{table}[H]
\captionsetup{font=small}
\caption{\textbf{In-Palm Rotation teacher reward terms.} Weights $w_k$ are listed in the last row.}
\label{table:reward_inpalm}
\vspace{-2mm}
\renewcommand{\arraystretch}{1.4}
\begin{tabularx}{\linewidth}{@{}>{\raggedleft\arraybackslash}p{0.34\linewidth}@{\hspace{3pt}}|@{\hspace{3pt}}>{\arraybackslash}X@{}}
    \Xhline{1\arrayrulewidth}
    \textbf{Reward Term} & \textbf{Expression} \\
    \hline
    \multicolumn{2}{c}{\textit{\textbf{Task Rewards}}: \raisebox{0.1ex}{$\mathcal{R}_{\text{task}} = \sum_{k=1}^{4} r_k$}} \\
    \Xhline{1\arrayrulewidth}
    \text{Goal Reached} ($r_1$) & \raisebox{0.1ex}{$w_1 \; \mathds{1}_{\theta_t < 0.1}$} \\
    \hline
    \text{Rotation Progress} ($r_2$) & \raisebox{0.1ex}{$w_2 \; \min(\Delta\theta_t,\, 6)$} \\
    \hline
    \text{Fingertip Proximity} ($r_3$) & \raisebox{0.1ex}{$w_3 \; \exp(-d_{\min,t}^2 / 0.04^2)$} \\
    \hline
    \text{Drop} ($r_4$) & \raisebox{0.1ex}{$w_4 \; \mathds{1}_{\text{dropped}}$} \\
    \Xhline{1\arrayrulewidth}
    \multicolumn{2}{c}{\textit{\textbf{Regularization}}: \raisebox{0.1ex}{$\mathcal{R}_{\text{reg}} = \sum_{k=5}^{10} r_k$}} \\
    \Xhline{1\arrayrulewidth}
    \text{Off-Axis Orientation} ($r_5$) & \raisebox{0.1ex}{$w_5 \; \theta^{\perp}_t$} \\
    \hline
    \text{Object Displacement} ($r_6$) & \raisebox{0.1ex}{$w_6 \; [\norm{\bp_t - \bp^{\ast}}_2 - 0.02]_+$} \\
    \hline
    \text{Contact Force} ($r_7$) & \raisebox{0.1ex}{$w_7 \; \min([\norm{\b{f}_t}_2 - 1]_+,\, 20)$} \\
    \hline
    \text{Action Rate} ($r_8$) & \raisebox{0.1ex}{$w_8 \; \norm{\ba_t - \ba_{t-1}}_2^2$} \\
    \hline
    \text{Joint Limits} ($r_9$) & \raisebox{0.1ex}{$w_9 \; \sum_j \big([q_{j,t} - q_j^{\max}]_+ + [q_j^{\min} - q_{j,t}]_+\big)$} \\
    \hline
    \text{Work} ($r_{10}$) & \raisebox{0.1ex}{$w_{10} \; \sum_j |\tau_{j,t}\, \dot{q}_{j,t}|$} \\
    \Xhline{1\arrayrulewidth}
    \multicolumn{2}{@{}p{\linewidth}@{}}{\textbf{Weights}: $w_1{=}150$, $w_2{=}2$, $w_3{=}0.1$, $w_4{=}-100$, $w_5{=}-1$, $w_6{=}-100$, $w_7{=}-5$, $w_8{=}-5{\times}10^{-4}$, $w_9{=}-40$, $w_{10}{=}-5{\times}10^{-3}$} \\
    \Xhline{1\arrayrulewidth}
\end{tabularx}
\end{table}

\begin{table}[H]
\captionsetup{font=small}
\caption{\textbf{Grasp-and-Lift teacher reward terms.} Sharpa and Allegro share the same terms and weights.}
\label{table:reward_grasp}
\vspace{-2mm}
\renewcommand{\arraystretch}{1.4}
\begin{tabularx}{\linewidth}{@{}>{\raggedleft\arraybackslash}p{0.34\linewidth}@{\hspace{3pt}}|@{\hspace{3pt}}>{\arraybackslash}X@{}}
    \Xhline{1\arrayrulewidth}
    \textbf{Reward Term} & \textbf{Expression} \\
    \hline
    \multicolumn{2}{c}{\textit{\textbf{Task Rewards}}: \raisebox{0.1ex}{$\mathcal{R}_{\text{task}} = \sum_{k=1}^{3} r_k$}} \\
    \Xhline{1\arrayrulewidth}
    \text{Sustained Lift} ($r_1$) & \raisebox{0.1ex}{$w_1 \; \big(h_t + 10 \; \mathds{1}_{h_t > 0.05}\big)$} \\
    \hline
    \text{Goal Tracking} ($r_2$) & \raisebox{0.1ex}{$w_2 \; \exp(-\norm{\bp_t - \bp^{\text{goal}}}_2^2 / 0.25^2)$} \\
    \hline
    \text{Drop} ($r_3$) & \raisebox{0.1ex}{$w_3 \; \mathds{1}_{\text{dropped}}$} \\
    \Xhline{1\arrayrulewidth}
    \multicolumn{2}{c}{\textit{\textbf{Regularization}}: \raisebox{0.1ex}{$\mathcal{R}_{\text{reg}} = \sum_{k=4}^{6} r_k$}} \\
    \Xhline{1\arrayrulewidth}
    \text{Fingertip Proximity} ($r_4$) & \raisebox{0.1ex}{$w_4 \; \tfrac{1}{N}\sum_{i=1}^{N_\text{finger}} \exp(-d_{i,t}^2 / 0.03^2)$} \\
    \hline
    \text{Palm Orientation} ($r_5$) & \raisebox{0.1ex}{$w_5 \; [-\hat{\bn}_t \cdot \hat{\bz}]_+$} \\
    \hline
    \text{Action Rate} ($r_6$) & \raisebox{0.1ex}{$w_6 \; \norm{\ba_t - \ba_{t-1}}_2^2$} \\
    \Xhline{1\arrayrulewidth}
    \multicolumn{2}{@{}p{\linewidth}@{}}{\textbf{Weights}: $w_1{=}5$, $w_2{=}1$, $w_3{=}-30$, $w_4{=}1$, $w_5{=}-0.5$, $w_6{=}-10^{-3}$} \\
    \Xhline{1\arrayrulewidth}
\end{tabularx}
\end{table}

\begin{table}[H]
\captionsetup{font=small}
\caption{\textbf{Teacher and student training settings.} Paired values are listed as In-Palm Rotation / Grasp-and-Lift.}
\label{table:training}
\vspace{-2mm}
\renewcommand{\arraystretch}{1.3}
\begin{tabularx}{\linewidth}{@{}>{\raggedleft\arraybackslash}p{0.34\linewidth}@{\hspace{3pt}}|@{\hspace{3pt}}>{\arraybackslash}X@{}}
    \Xhline{1\arrayrulewidth}
    \textbf{Setting} & \textbf{Value} \\
    \hline
    \multicolumn{2}{c}{\textit{\textbf{Simulation}}} \\
    \Xhline{1\arrayrulewidth}
    Physics step & 5\,ms \\
    \hline
    Control step & 25\,ms \\
    \hline
    Episode length & 10\,s / 20\,s \\
    \Xhline{1\arrayrulewidth}
    \multicolumn{2}{c}{\textit{\textbf{Teacher (PPO)}}} \\
    \Xhline{1\arrayrulewidth}
    Parallel environments & 8192 / 4096 \\
    \hline
    Steps per environment & 24 \\
    \hline
    Iterations & 20000 / 5000 \\
    \hline
    Epochs, mini-batches & 5, 4 \\
    \hline
    Learning rate & $10^{-3}$, adaptive (target KL $0.01$) \\
    \hline
    $\gamma$, $\lambda$, clip & $0.998$, $0.95$, $0.2$ \\
    \hline
    Entropy coefficient & $0.002$ / $0.001$ \\
    \hline
    Gradient clipping & $1.0$ \\
    \hline
    Actor, critic & MLP 512-256-128, ELU \\
    \Xhline{1\arrayrulewidth}
    \multicolumn{2}{c}{\textit{\textbf{Student (Distillation)}}} \\
    \Xhline{1\arrayrulewidth}
    Scheme & DAgger, $\beta{=}0$ \\
    \hline
    Loss & MSE on clipped actions \\
    \hline
    Parallel environments & 2048 \\
    \hline
    Steps per environment & 24 \\
    
    \hline
    Auxiliary losses & object size, goal angle (MSE, weight $1.0$) / none \\
    \hline
    Iterations & 3000 \\
    \hline
    Optimizer & Adam, $10^{-4}$, 5 epochs \\
    \hline
    Policy head & MLP 512-256-128, ELU; fixed std $1.0$ \\
    \Xhline{1\arrayrulewidth}
\end{tabularx}
\end{table}

\FloatBarrier
\subsection{Tactile Encoders} 
\label{app:encoders}
\begin{table}[t]
\captionsetup{font=small}
\caption{\textbf{Tactile encoders.} The spatial module runs per link block. Each encoder ends in FC(32), and the policy head takes its output together with the 64-D proprioceptive feature.}
\label{table:encoder}
\vspace{-2mm}
\centering
\footnotesize
\setlength{\tabcolsep}{3pt}
\begin{tabularx}{\columnwidth}{@{}l>{\raggedright\arraybackslash}Xl@{}}
\toprule
\rowcolor{refrow} \multicolumn{3}{@{}l}{\textit{Tactile encoders}} \\
Encoder & Spatial module & Temporal \\
\addlinespace[2pt]
LSTM          & --                                             & LSTM(64) \\
CNN           & Conv3$\times$3(16)--LN--Conv3$\times$3(32)--LN & GRU(64) \\
Link-wise MLP & FC(64)--FC(64), shared, ELU                    & GRU(64) \\
\addlinespace[2pt]
\midrule
\rowcolor{refrow} \multicolumn{3}{@{}l}{\textit{Shared}} \\
Proprioceptive encoder & \multicolumn{2}{l}{LSTM(128)--FC(64)} \\
Policy head            & \multicolumn{2}{l}{FC(512)--FC(256)--FC(128), ELU} \\
\bottomrule
\end{tabularx}
\vspace{-4mm}
\end{table}

Table~\ref{table:encoder} summarizes the three tactile encoders and the shared modules. The input is the per-link block of Section III.E, and each cell holds the measurement $x_{i,t}$ of Section III.D as six channels.
The LSTM encoder flattens the cells of all blocks into one input vector.
The CNN encoder applies a separate CNN to each block. The convolutions preserve the spatial size. The output of each block is flattened, reduced by a per-link linear layer, and concatenated over links into a 32-D vector.
The link-wise MLP encoder applies the same MLP to every cell of a block
and averages the outputs over all cells of the block. The per-link
averages are averaged over links and mapped to a 32-D vector by a
linear layer.                                              %
Because the average is invariant to the assignment of taxels to cells,
the three layouts of Section~\ref{sec:representation} are the same
input to this encoder, and the spread of its three cells in
Table~\ref{tab:representation} reflects the seed variance of this encoder.

\subsection{Placement Details}
\label{app:placement}

All task-informed methods use the contact statistics of $\pi_T$. For \taskname{In-Palm Rotation}, \methodname{Top-contact}, \methodname{Contact-cover}, and \methodname{Least-contact} assign at least one taxel to every link used in the task.
\methodname{Contact-cover} is computed by lazy greedy selection.
Contact events are evaluated on every fourth timestep.
\methodname{Least-contact} excludes the 879 candidates with $|\mathcal{E}_i| = 0$.
Candidates with equal $|\mathcal{E}_i|$ are ordered by their index.
At $B{=}50$, this ordering determines 23 of the 50 selected taxels.
The magnitude of \methodname{Per-link force} is bounded between $10^{-4}$\,N and 10\,N.
Gaussian noise with a standard deviation of 0.08\,N is then added.
The resulting observation lies in $\mathbb{R}^{18}$.

\subsection{Metrics and Evaluation Protocol}
\label{app:metric}

All instances of $\pi_S$ are evaluated at the same training iteration, using the mean of the action distribution.
We report the mean and standard deviation over three seeds.
Only the first episode of each environment is counted.
The episode length and the domain randomization are the same as in training.
$\pi_T$ is evaluated under the same protocol.

For \taskname{In-Palm Rotation}, goals are successive quarter turns about a fixed horizontal axis, starting from the initial object pose.
A goal is reached when the error about the rotation axis is below 0.1\,rad and the off-axis error is below $\pi/6$, after which the next goal is set.
An episode terminates when the object falls. An episode is a success if at least one goal is reached before the fall.

For \taskname{Grasp-and-Lift}, the object height is measured relative to its height at reset.
Measurement begins once the object is lifted above a fixed threshold.
The held fraction is the fraction of subsequent timesteps at which the object remains above this threshold.
Episodes in which the object is never lifted are excluded from the held fraction.

A disturbance is a wrench applied to the center of mass of the object.
Each pulse is a force, a force with a lever offset, or a pure torque, chosen with equal probability, and its direction is a random unit vector.
The strength coefficient is 0.25, which corresponds to forces of 1 to 2\,N and torques of 0.02 to 0.035\,N\,m, independent of the object mass.
Each pulse follows a half-sine profile lasting 0.1 to 0.5\,s, and the next pulse starts 2 to 4\,s after it ends.
The first pulse is applied after 1\,s of the episode, once the object has been lifted 5\,cm above its reset height.
Training leaves 25\% of the episodes without disturbances, whereas evaluation applies them in every episode.
Both hands use the same settings.
\clearpage
\raggedbottom

\subsection{Supplementary Analyses}
\label{app:analysis}

\subsubsection{Contact Coverage}
\tabref{tab:contact_events} shows how much of the contact of $\pi_T$ each placement method observes.
Task-informed methods detect most contact events at the smallest budget, whereas \methodname{Uniform} requires more than 100 taxels to reach the same level.
For the task-informed methods, detection saturates at intermediate budgets while the captured readings continue to rise.
Additional taxels therefore observe the detected contacts in more detail.
\methodname{Contact-cover} detects all contact events at $B{=}300$ but captures about half of the readings, since taxels added after full coverage follow index order (Appendix~\ref{app:placement}).
Redundancy rises with the budget for \methodname{Uniform}, whereas \methodname{Top-contact} is redundant from the smallest budget.
For \methodname{Top-contact}, the redundant taxels lie at the most contacted sites, so its captured readings remain the highest at every budget.
\begin{table}[t!]
  \centering
  \footnotesize
  \setlength{\tabcolsep}{5pt}
  \renewcommand{\arraystretch}{0.75}
  \captionsetup{font=footnotesize}
\caption{\textbf{Contact events, taxel readings, and redundancy of each placement method.}~\taskname{In-Palm Rotation} on the XHand1. \textit{Contact Events} is the fraction of contact events at which at least one selected taxel on the contacted link reports a non-zero reading. \textit{Taxel Readings} is the fraction of all non-zero readings in the rollouts of $\pi_T$ that fall on the selected taxels. \textit{Redundancy} is the fraction of a taxel's contacts that its nearest selected neighbor also detects.}
\label{tab:contact_events}
  \vspace{-1mm}
  \begin{tabular}{l r rrr}
  \toprule
  & \multicolumn{1}{c}{\scriptsize Task-agnostic} & \multicolumn{3}{c}{\scriptsize Task-informed} \\
  \cmidrule(lr){2-2} \cmidrule(lr){3-5}
  \rs Budget ($B$) & Uniform & Top-contact & Contact-cover & Link-prop. \\
  \specialrule{0.4pt}{0pt}{1.2pt}
  \specialrule{0.4pt}{0pt}{0pt}
  \rowcolor{refrow}\multicolumn{5}{l}{\rs\textit{Contact Events (\%)}} \\
  \rs \pz10  & 0.1 & 83.9 & \pz88.8   & \pz0.1 \\
  \rs \pz15  & \pz0.2 & 84.7 & \pz94.8   & \pz0.1 \\
  \rs \pz25  & 24.6   & 85.6 & \pz98.1   & 25.1   \\
  \rs \pz50  & 35.9   & 87.5 & \pz99.7   & 54.4   \\
  \rs \pz75  & 45.8   & 88.9 & \pz99.9   & 59.5   \\
  \rs 100    & 72.3   & 93.7 & 99.9   & 64.4   \\
  \rs 300    & 91.8   & 99.8 & 100.0     & 84.5   \\
  \midrule
  \rowcolor{refrow}\multicolumn{5}{l}{\rs\textit{Taxel Readings (\%)}} \\
  \rs \pz10  & \pz0.0 & \pz8.1 & \pz5.1 & \pz0.0 \\
  \rs \pz15  &0.1 & 14.5   & \pz7.6 & 0.1 \\
  \rs \pz25  & \pz0.9 & 26.4   & 12.7   & \pz0.9 \\
  \rs \pz50  & \pz1.7 & 48.7   & 18.3   & \pz2.7 \\
  \rs \pz75  & \pz3.4 & 63.1   & 26.9   & \pz4.5 \\
  \rs 100    & \pz5.5 & 72.8   & 31.7   & \pz7.9 \\
  \rs 300    & 14.4   & 98.0   & 52.6   & 17.4   \\
  \midrule
  \rowcolor{refrow}\multicolumn{5}{l}{\rs\textit{Redundancy (\%)}} \\
  \rs \pz25  & \pz8.4 & 95.7 & 76.1 & 11.1 \\
  \rs 100    & 25.2   & 91.3 & 75.9 & 29.7 \\
  \rs 300    & 35.0   & 79.7 & 76.6 & 44.0 \\
  \bottomrule
  \end{tabular}
  \vstab
\end{table}

\subsubsection{Episode Outcomes}
\tabref{tab:failure} shows how the episodes of \taskname{In-Palm Rotation} end.
The drop rate is similar with and without tactile input.
Only \methodname{Uniform} and \methodname{Link-proportional} at the smallest budget drop the object more often, and these two detect few contact events at that budget (\tabref{tab:contact_events}).
Tactile input reduces under-rotation, and the reduction grows with the budget.
In this task, touch therefore serves to advance the rotation rather than to hold the object.

\begin{table}[H]
  \centering
  \footnotesize
  \setlength{\tabcolsep}{3pt}
  \renewcommand{\arraystretch}{0.75}
  \captionsetup{font=footnotesize}
  \caption{\textbf{Analysis of \taskname{In-Palm Rotation} episodes across placement methods and budgets (\%).}~XHand1 with the LSTM encoder. Each episode is classified as success, drop, or under-rotation. Drop denotes a fall before the first goal, and under-rotation denotes an episode that ends without a goal or a fall.}
  \label{tab:failure}
  \vspace{-1mm}
  \begin{tabular}{r l r@{$\pm$}l r@{$\pm$}l r@{$\pm$}l}
  \toprule
  \multicolumn{1}{l}{\rs $B$} & Placement & \multicolumn{2}{c}{Success} & \multicolumn{2}{c}{Drop} & \multicolumn{2}{c}{Under-rotation} \\
  \specialrule{0.4pt}{0pt}{1.2pt}
  \specialrule{0.4pt}{0pt}{0pt}
  \rs 0    & No tactile    &  0.0 & \pz0.0 & 15.0 & 12.7   & 85.0 & 12.7   \\
  \specialrule{0.4pt}{0pt}{1.2pt}
  \specialrule{0.4pt}{0pt}{0pt}
  \rs 25   & \methodname{Uniform}       & 12.4 & \pz1.3 & 30.8 & \pz5.1 & 56.8 & \pz4.5 \\
  \rs      & \methodname{Top-contact}   & 50.5 & \pz1.3 & 16.2 & \pz1.1 & 33.3 & \pz3.2 \\
  \rs      & \methodname{Contact-cover} & 50.0 & \pz1.8 & 14.6 & \pz3.6 & 35.4 & \pz4.5 \\
  \rs      & \methodname{Link-prop.}    &  8.6 & \pz3.1 & 30.2 & \pz5.3 & 61.2 & \pz4.7 \\
  \midrule
  \rs 100  & \methodname{Uniform}       & 51.4 & \pz2.3 & 16.5 & \pz3.2 & 32.1 & \pz1.1 \\
  \rs      & \methodname{Top-contact}   & 56.6 & \pz2.9 & 14.1 & \pz1.0 & 29.3 & \pz0.3 \\
  \rs      & \methodname{Contact-cover} & 55.7 & \pz3.8 & 16.8 & \pz1.5 & 27.5 & \pz4.3 \\
  \rs      & \methodname{Link-prop.}    & 48.8 & \pz0.9 & 18.2 & \pz1.7 & 33.0 & \pz3.3 \\
  \midrule
  \rs 300  & \methodname{Uniform}       & 56.2 & \pz3.6 & 13.9 & \pz2.0 & 29.9 & \pz0.1 \\
  \rs      & \methodname{Top-contact}   & 58.0 & \pz2.8 & 13.8 & \pz0.8 & 28.2 & \pz3.2 \\
  \rs      & \methodname{Contact-cover} & 59.4 & \pz2.8 & 15.3 & \pz0.6 & 25.3 & \pz1.9 \\
  \rs      & \methodname{Link-prop.}    & 56.9 & \pz1.3 & 15.4 & \pz1.2 & 27.7 & \pz3.9 \\
  \specialrule{0.4pt}{0pt}{1.2pt}
  \specialrule{0.4pt}{0pt}{0pt}
  \rs 2000 & All           & 61.5 & \pz0.3 & 13.1 & \pz1.3 & 25.4 & \pz2.8 \\
  \bottomrule
  \end{tabular}
  \vstab
\end{table}

\subsubsection{Contact Concentration}
\begin{table}[H]
  \centering
  \footnotesize
  \setlength{\tabcolsep}{5pt}
  \renewcommand{\arraystretch}{0.75}
  \captionsetup{font=footnotesize}
  \caption{\textbf{Contact concentration per task and hand.}~Rollouts of $\pi_T$ on the 2000 candidate taxels of each hand. Top-100 Share: ratio of contact readings on the 100 most contacted taxels. Entropy: Shannon entropy over taxels, normalized by $\log 2000$. Links Spanned: links holding a top-100 taxel over links with candidate taxels.}
  \label{tab:contact_dist}
  \vspace{-1mm}
  \begin{tabular}{l r r r}
  \toprule
  \rs Hand & Top-100 Share (\%) & Entropy & Links Spanned \\
  \specialrule{0.4pt}{0pt}{1.2pt}
  \specialrule{0.4pt}{0pt}{0pt}
  \rowcolor{refrow}\multicolumn{4}{l}{\rs\taskname{In-Palm Rotation}} \\
  \rs \quad XHand1  & 73.5 & 0.70 & \pz3 / \pz6 \\
  \midrule
  \rowcolor{refrow}\multicolumn{4}{l}{\rs\taskname{Grasp-and-Lift}} \\
  \rs \quad Allegro & 66.0 & 0.72 & 16 / 39 \\
  \rs \quad Sharpa  & 38.3 & 0.84 & \pz5 / 17 \\
  \bottomrule
  \end{tabular}
  \vstab
\end{table}

\tabref{tab:contact_dist} compares how the contacts of $\pi_T$ concentrate on each hand.
On the XHand1 and the Allegro, the 100 most contacted taxels account for most of the contact readings, whereas on the Sharpa they account for less than half.
The Allegro is as concentrated as the XHand1, but its top taxels span four fingers rather than the palm and the thumb.

\end{document}